\documentclass[journal]{IEEEtran}

\usepackage{cite}
\usepackage{graphicx}
\usepackage{amsmath,amssymb}
\usepackage{algorithm}
\usepackage{algorithmic}
\usepackage{multirow}
\usepackage{url}
\usepackage{hyperref}       
\usepackage{booktabs}   
\usepackage{siunitx}
\usepackage{tikz}
\usetikzlibrary{arrows.meta, positioning, calc, matrix}

\newcommand{\fedskiptwin}{\textsc{FedSkipTwin}}
\newcommand{\fedavg}{\textsc{FedAvg}}

\begin{document}

\title{FedSkipTwin: Digital-Twin-Guided Client Skipping for Communication-Efficient Federated Learning}

\author{Daniel Commey, Kamel Abbad, Garth V. Crosby, and Lyes Khoukhi%
\thanks{Daniel Commey and Garth V. Crosby are with the Department of Multidisciplinary Engineering, Texas A\&M University, College Station, TX, USA (e-mail: dcommey@tamu.edu, gvcrosby@tamu.edu).}%
\thanks{Kamel Abbad is with GREYC Laboratory, NEAC Industry, University of Caen Normandy, Caen, France (e-mail: kamel.abbad@unicaen.fr).}%
\thanks{Lyes Khoukhi is with ENSICAEN, Normandie University, Caen, France (e-mail: lyes.khoukhi@ensicaen.fr).}%
}

\maketitle

\begin{abstract}
Communication overhead remains a primary bottleneck in federated learning (FL), particularly for applications involving mobile and IoT devices with constrained bandwidth. This work introduces \fedskiptwin, a novel client-skipping algorithm driven by lightweight, server-side \emph{digital twins}. Each twin, implemented as a simple LSTM, observes a client's historical sequence of gradient norms to forecast both the magnitude and the epistemic uncertainty of its next update. The server leverages these predictions, requesting communication only when either value exceeds a predefined threshold; otherwise, it instructs the client to skip the round, thereby saving bandwidth. Experiments are conducted on the UCI-HAR and MNIST datasets with 10 clients under a non-IID data distribution. The results demonstrate that \fedskiptwin\ reduces total communication by \textbf{12--15.5\%} across 20 rounds while simultaneously \emph{improving} final model accuracy by up to 0.5 percentage points compared to the standard \fedavg\ algorithm. These findings establish that prediction-guided skipping is a practical and effective strategy for resource-aware FL in bandwidth-constrained edge environments.
\end{abstract}


\section{Introduction}

Federated learning (FL) enables collaborative training of machine learning models across decentralized data sources—such as mobile phones and IoT devices—without sharing raw data \cite{mcmahan_communication-efficient_2017}. This paradigm offers strong privacy guarantees but suffers from a critical bottleneck: communication overhead. In each round of FL, model updates are exchanged between the central server and numerous clients. For large models or networks with limited bandwidth, this overhead can become a major constraint, limiting scalability and energy efficiency.

Existing work has largely addressed this problem by reducing the size of model updates through techniques such as gradient sparsification \cite{lin_deep_2020, han_adaptive_2020} and quantization \cite{reisizadeh_fedpaq_2020, alistarh_qsgd_2017}. However, these methods typically assume that all selected clients must communicate in every round, regardless of the significance of their updates. This raises a fundamental research question: \emph{Do all clients need to send updates in every round?}

We argue that they do not. In many cases, a client’s local update may be small or redundant, especially in later training stages or when the local data is poorly aligned with the current global model. Skipping such updates could yield communication savings with minimal impact on model performance.

To this end, we propose \fedskiptwin, a novel communication-efficient FL algorithm that leverages lightweight \emph{server-side digital twins} to selectively skip clients during training. Each digital twin is a compact LSTM-based model that learns to forecast the magnitude and uncertainty of its associated client’s gradient updates over time. Before each round, the server uses these forecasts to determine whether a client is likely to produce a significant update. Only clients with high predicted impact or high uncertainty are asked to participate, while the rest are safely skipped.

Our approach introduces no additional burden on client devices, as all intelligence and forecasting are handled by the server. Furthermore, the decision to skip is governed by a conservative \emph{dual-threshold mechanism}, ensuring that only confidently low-impact updates are skipped, thereby preserving convergence integrity.

This paper makes the following key contributions:
\begin{itemize}
    \item We introduce \textbf{digital twins} for FL: lightweight server-side LSTM models that predict both the magnitude and uncertainty of future client updates.
    \item We design a \textbf{dual-threshold skip rule} that uses these predictions to conservatively skip unnecessary client communications, saving bandwidth without degrading model accuracy.
    \item We implement \fedskiptwin\ using the Flower framework \cite{beutel_flower_2022} and evaluate it on two benchmarks (UCI-HAR and MNIST) under non-IID data distributions.
    \item We show that \fedskiptwin\ reduces total communication by up to \textbf{15.5\%} while slightly \emph{improving} final model accuracy compared to \fedavg.
\end{itemize}

\textbf{Structure of the Paper:}
Section~\ref{sec:related_work} reviews related work on communication-efficient FL and the emerging role of digital twins. Section~\ref{sec:methodology} details the design of the \fedskiptwin\ algorithm, including the digital twin architecture and skip decision mechanism. Section~\ref{sec:experimental_setup} outlines the experimental setup, datasets, and configurations. Section~\ref{sec:results} presents results comparing \fedskiptwin\ with \fedavg\ in terms of accuracy and communication efficiency. Section~\ref{sec:discussion} discusses key insights, limitations, and future directions. Section~\ref{sec:conclusion} concludes the paper.

\section{Related Work}
\label{sec:related_work}

\subsection{Communication-Efficient Federated Learning}
The challenge of communication overhead in FL has been a primary driver of research in the field. Broadly, existing strategies fall into several categories.

\textbf{Gradient Compression} methods aim to reduce the size of each model update message. This includes \textit{sparsification}, which transmits only the most significant gradient values \cite{lin_deep_2020, han_adaptive_2020}, and \textit{quantization}, which reduces the numerical precision of the gradients, for instance by converting 32-bit floats to 8-bit integers or less \cite{reisizadeh_fedpaq_2020, alistarh_qsgd_2017}. While effective, these methods often introduce information loss and still require a transmission in every round.

\textbf{Client Selection} protocols reduce communication frequency by only involving a subset of clients in each round. Selection can be based on system resources \cite{nishio_client_2019}, data quality, or local loss values \cite{cho_client_2020}. These methods differ from our work in that they typically react to the current state of clients rather than proactively forecasting the future utility of a client's update. \fedskiptwin\ could be used in conjunction with these methods, first selecting a pool of available clients and then using twins to decide which ones should actually compute and send an update.

\textbf{Local Update Strategies}, most famously embodied by \fedavg\ \cite{mcmahan_communication-efficient_2017}, reduce the number of communication rounds by allowing clients to perform multiple local training epochs before sending an update. This amortizes the communication cost over more computation but does not change the fact that all selected clients must communicate at the end of the round.

Beyond direct communication reduction, the overall trustworthiness of FL depends on addressing security and privacy vulnerabilities. These challenges are orthogonal but equally important for real-world deployment. For instance, ensuring the integrity of the collaborative process is crucial, and recent work has explored using zero-knowledge proofs to enable verifiable and privacy-preserving model evaluation \cite{commey_zkp-fedeval_2025}. The communication-saving techniques proposed in \fedskiptwin\ are complementary to such security measures and could be combined to build systems that are not only efficient but also private and robust.

\subsection{Digital Twins in Machine Learning}
The concept of a digital twin (a virtual model of a physical asset or system) was pioneered in manufacturing and industrial IoT for monitoring, diagnostics, and simulation. It allows for virtual testing and optimization before applying changes to the physical world \cite{grieves_digital_2014}. While its application in core machine learning is nascent, the idea of using a lightweight surrogate model to guide a more complex process is powerful. To our knowledge, this work is the first to apply the digital twin concept to model the behavior of individual clients in an FL system for the purpose of optimizing communication scheduling.

\section{Methodology}
\label{sec:methodology}

\subsection{Problem Formulation}
Consider a standard FL system with a central server and a set of $N$ clients, indexed by $i \in \{1, \dots, N\}$. Each client $i$ possesses a local dataset $D_i$ that is not shared with the server or other clients. The global learning objective is to find the model parameters $\theta$ that minimize a global loss function $F(\theta)$, defined as a weighted average of the local loss functions $F_i(\theta)$:
\begin{equation}
\min_{\theta} F(\theta) = \sum_{i=1}^{N} \frac{|D_i|}{|D|} F_i(\theta)
\end{equation}
where $|D| = \sum_{i=1}^{N} |D_i|$ is the total size of the data across all clients. The \fedavg\ algorithm \cite{mcmahan_communication-efficient_2017} addresses this by iteratively distributing the global model $\theta_t$ to clients, having them compute local updates, and aggregating these updates to form the next global model $\theta_{t+1}$.

\subsection{The FedSkipTwin Algorithm}
\begin{figure}[!htbp]
\centering

\begin{tikzpicture}[
    every node/.style = {font=\footnotesize},
    arr/.style  = {-{Latex[length=2mm]}, thick},
    skip/.style = {arr, dashed, red},
    srv/.style  = {draw, rounded corners, fill=blue!5,
                   minimum width=2.8cm, minimum height=1.0cm,
                   align=center},
    twin/.style = {draw, rounded corners=2pt, fill=gray!15,
                   minimum width=2.2cm, minimum height=0.8cm,
                   align=center, font=\scriptsize},
    cli/.style  = {draw, rounded corners, fill=green!5,
                   minimum width=2.2cm, minimum height=0.9cm,
                   align=center},
    dec/.style  = {draw, rounded corners, fill=yellow!15,
                   minimum width=2.8cm, minimum height=0.9cm,
                   align=center}
  ]

\matrix[column sep=0.9cm, row sep=0.7cm]{
  \node[srv]  (S) {FL \textbf{Server}\\Global model $\theta_t$}; &
  \node[twin] (T1) {Twin$_1$\\\scriptsize LSTM}; \\

  \node (empty) {}; & 
  \node[twin] (T2) {Twin$_2$\\\scriptsize LSTM}; \\

  \node (empty2) {}; &
  \node[twin] (T3) {Twin$_N$\\\scriptsize LSTM}; \\

  \node[dec] (D) {Dual-threshold\\skip rule}; &
  \node[cli] (C1) {Client 1\\Data $D_1$}; \\

  \node (empty3) {}; &
  \node[cli] (C2) {Client 2\\Data $D_2$}; \\

  \node (empty4) {}; &
  \node[cli] (CN) {Client $N$\\Data $D_N$}; \\
};

\foreach \t in {T1,T2,T3}{
  \draw[arr, gray] (S.east) -- ++(0.15,0) |- (\t.west);
}

\draw[arr] (S.south) -- (D.north)
            node[midway, left, xshift=-2pt, font=\tiny, align=right]
            {predict\\$\lVert g_i\rVert_2$, uncertainty};

\draw[arr]  (D.east) -- (C1.west) node[pos=.35, above, font=\tiny] {communicate};
\draw[skip] (D.east) -- (C2.west) node[pos=.35, above, font=\tiny, red] {skip};
\draw[arr]  (D.east) -- (CN.west) node[pos=.35, above, font=\tiny] {communicate};

\draw[arr, blue!70!black] (C1.east) -- ++(0.4,0) |- (T1.east)
      node[midway, above, font=\tiny] {actual norm};
\draw[arr, blue!70!black] (CN.east) -- ++(0.4,0) |- (T3.east);

\end{tikzpicture}

\caption{Overview of \fedskiptwin. The server hosts a lightweight digital twin for each client to forecast its next gradient‐norm and uncertainty. The dual-threshold rule decides whether a client should \emph{communicate} or \emph{skip}. Participating clients feed back their actual norms to retrain their twins.}
\label{fig:architecture}
\end{figure}
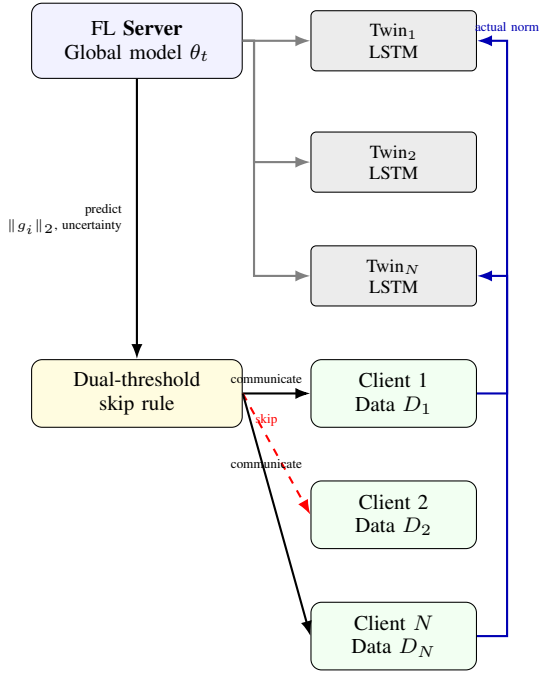

Our proposed method, \fedskiptwin, modifies this process by introducing a server-side digital twin for each client. As depicted in Figure~\ref{fig:architecture}, each twin, $\text{Twin}_i$, is an LSTM-based time-series model. Its purpose is to learn the temporal dynamics of client $i$'s update significance, which we proxy using the L2 norm of the client's gradient update vector.

The algorithm proceeds in rounds. At the start of round $t$, for each client $i$, the server uses $\text{Twin}_i$ to predict the L2 norm of the client's anticipated update, $||\mathbf{g}_i^{(t)}||_2$, based on a sequence of its previously observed norms. The twin also estimates its own epistemic uncertainty for this prediction. This is achieved by using Monte Carlo dropout during the twin's inference step \cite{gal_dropout_2016}, where multiple stochastic forward passes yield a distribution of predictions, whose variance reflects model uncertainty.

The server then applies the skip decision logic detailed below. Clients instructed to train proceed as in \fedavg, while skipped clients do nothing, saving both computation and communication. The full process is outlined in Algorithm~\ref{alg:fedskiptwin}.

\begin{algorithm}[!htbp]
\caption{\fedskiptwin\ Algorithm}
\label{alg:fedskiptwin}
\begin{algorithmic}[1]
\STATE \textbf{Input:} Initial model $\theta_0$, skip thresholds $\tau_{mag}, \tau_{unc}$, local epochs $E$, rounds $T$
\STATE \textbf{Server Initializes:} $\text{Twin}_i$ for all clients $i=1, \dots, N$
\FOR{round $t = 1, 2, \ldots, T$}
    \STATE Server broadcasts current global model $\theta_{t-1}$ to clients
    \STATE $S_t \leftarrow \emptyset$ (set of participating clients)
    \FOR{each client $i$ in parallel}
        \STATE $(\text{pred\_mag}_i, \text{uncertainty}_i) = \text{Twin}_i.\text{predict()}$
        \IF{$\text{pred\_mag}_i \ge \tau_{mag}$ \textbf{or} $\text{uncertainty}_i \ge \tau_{unc}$}
            \STATE Server instructs client $i$ to train
            \STATE $\theta_i^{(t)} \leftarrow \text{ClientUpdate}(i, \theta_{t-1})$
            \STATE Client $i$ sends $\Delta_i^{(t)} = \theta_i^{(t)} - \theta_{t-1}$ to server
            \STATE $S_t \leftarrow S_t \cup \{i\}$
        \ELSE
            \STATE Server instructs client $i$ to skip
        \ENDIF
    \ENDFOR
    \STATE Server aggregates updates from participating clients:
    \STATE $\theta_t \leftarrow \theta_{t-1} + \sum_{i \in S_t} \frac{|D_i|}{\sum_{j \in S_t} |D_j|} \Delta_i^{(t)}$
    \FOR{each client $i \in S_t$}
        \STATE $\text{norm}_i^{(t)} = ||\Delta_i^{(t)}||_2$
        \STATE Server uses $\text{norm}_i^{(t)}$ to retrain $\text{Twin}_i$
    \ENDFOR
\ENDFOR
\end{algorithmic}
\end{algorithm}

\subsection{Skip Decision Mechanism}
The skip decision for client $i$ at round $t$ hinges on a dual-threshold check. A client is instructed to \textbf{skip} if and only if both its predicted gradient magnitude and the twin's prediction uncertainty are below their respective thresholds, $\tau_{mag}$ and $\tau_{unc}$:
\begin{equation}
\text{Decision}_i^{(t)} = 
\begin{cases} 
  \text{Skip} & \text{if } (\text{pred\_mag}_i < \tau_{mag}) \\
              & \quad \land (\text{uncertainty}_i < \tau_{unc}) \\
  \text{Communicate} & \text{otherwise}
\end{cases}
\end{equation}
This mechanism is deliberately conservative. It only skips a client when its twin is \textit{confident} that the update will be small. If the twin predicts a small update but is uncertain (i.e., high uncertainty), the client will still be asked to communicate. This design choice prioritizes model convergence integrity over maximizing communication savings, preventing the system from erroneously skipping potentially significant updates due to an inaccurate twin prediction.

\section{Experimental Setup}
\label{sec:experimental_setup}

\subsection{Datasets and Models}
We evaluate \fedskiptwin\ on two standard benchmarks for federated learning:
\begin{itemize}
    \item \textbf{UCI-HAR}: A human activity recognition dataset derived from smartphone sensor data, comprising 10,299 samples across 6 activities \cite{anguita_public_2013}. It is well-suited for simulating FL on edge devices.
    \item \textbf{MNIST}: The classic dataset of 70,000 handwritten digits, widely used for benchmarking image classification algorithms \cite{lecun_gradient-based_2002}.
\end{itemize}
The neural network architectures for each task are detailed in Table~\ref{tab:models}. These are standard, relatively simple models chosen to ensure our results are comparable to other works in the field.

\begin{table}[!htbp]
\centering
\caption{Neural network architectures used in experiments.}
\label{tab:models}
\begin{tabular}{@{}llp{4.5cm}@{}}
\toprule
\textbf{Dataset} & \textbf{Model} & \textbf{Layer Details} \\ \midrule
\multirow{3}{*}{UCI-HAR} & \multirow{3}{*}{MLP} & Dense (128 neurons, ReLU) \\
 & & Dense (64 neurons, ReLU) \\
 & & Dense (6 neurons, Softmax) \\ \midrule
\multirow{5}{*}{MNIST} & \multirow{5}{*}{CNN} & Conv2D (16 filters, 5x5, ReLU), MaxPool2D (2x2) \\
 & & Conv2D (32 filters, 5x5, ReLU), MaxPool2D (2x2) \\
 & & Flatten, Dense (10 neurons, Softmax) \\ \bottomrule
\end{tabular}
\end{table}

\subsection{Configuration}
Our experimental environment consists of a central server and 10 simulated clients, implemented using Python with PyTorch and the Flower federated learning framework \cite{beutel_flower_2022}. To simulate statistical heterogeneity, a common challenge in FL \cite{zhu_federated_2021}, we partitioned the data among the clients using a Dirichlet distribution with a concentration parameter of $\alpha=0.5$. This creates a non-IID (non-independent and identically distributed) setting where each client has a biased distribution of data labels.

Key hyperparameters are listed below:
\begin{itemize}
    \item \textbf{Clients}: 10 participants
    \item \textbf{Data distribution}: Non-IID (Dirichlet, $\alpha=0.5$)
    \item \textbf{Communication rounds}: 20
    \item \textbf{Local epochs ($E$)}: 3 per round
    \item \textbf{Batch size}: 32
    \item \textbf{Skip thresholds ($\tau_{mag}, \tau_{unc}$)}: $0.001$ (tuned via grid search)
\end{itemize}
All experiments were executed on a machine with a Ryzen 7 7700 CPU, 32 GB RAM, and an NVIDIA RTX 3060 Ti GPU.

\section{Results}
\label{sec:results}

We now present the empirical performance of \fedskiptwin\ against the \fedavg\ baseline, focusing on model accuracy, communication savings, and convergence dynamics.

\subsection{Performance Comparison}
Table~\ref{tab:results} summarizes the final model accuracy and total communication volume for both algorithms. On both datasets, \fedskiptwin\ not only reduces communication costs but also achieves slightly higher final test accuracy. On UCI-HAR, it cuts communication by \textbf{15.5\%} while improving accuracy by 0.5 percentage points. On MNIST, it achieves a \textbf{12.0\%} reduction in traffic with a marginal accuracy gain.

These results are significant because they show that intelligently skipping rounds does not compromise model quality. On the contrary, by filtering out updates that are predicted to be of low magnitude (which may represent noise or redundant information, especially in later training stages) \fedskiptwin\ may introduce a subtle regularization effect that leads to better generalization.

\begin{table}[!htbp]
\centering
\caption{Final accuracy and communication cost of \fedavg\ vs.\ \fedskiptwin\ after 20 rounds.}
\label{tab:results}
\small
\resizebox{\linewidth}{!}{%
\begin{tabular}{@{}lcccc@{}}
\toprule
\multirow{2}{*}{\textbf{Dataset}} & \multicolumn{2}{c}{\textbf{Accuracy}} & \multicolumn{2}{c}{\textbf{Communication (MB)}} \\
\cmidrule(lr){2-3} \cmidrule(lr){4-5}
 & \fedavg & \fedskiptwin & \fedavg & \fedskiptwin (\%) \\
\midrule
UCI-HAR & 0.9243 & \textbf{0.9291} & 135.45 & \textbf{114.46} (\SI{-15.5}{\percent}) \\
MNIST   & 0.9656 & \textbf{0.9669} & 408.80 & \textbf{359.75} (\SI{-12.0}{\percent}) \\
\bottomrule
\end{tabular}
}
\end{table}

\subsection{Convergence and Efficiency Analysis}
\begin{figure}[!htbp]
\centering
\includegraphics[width=\linewidth]{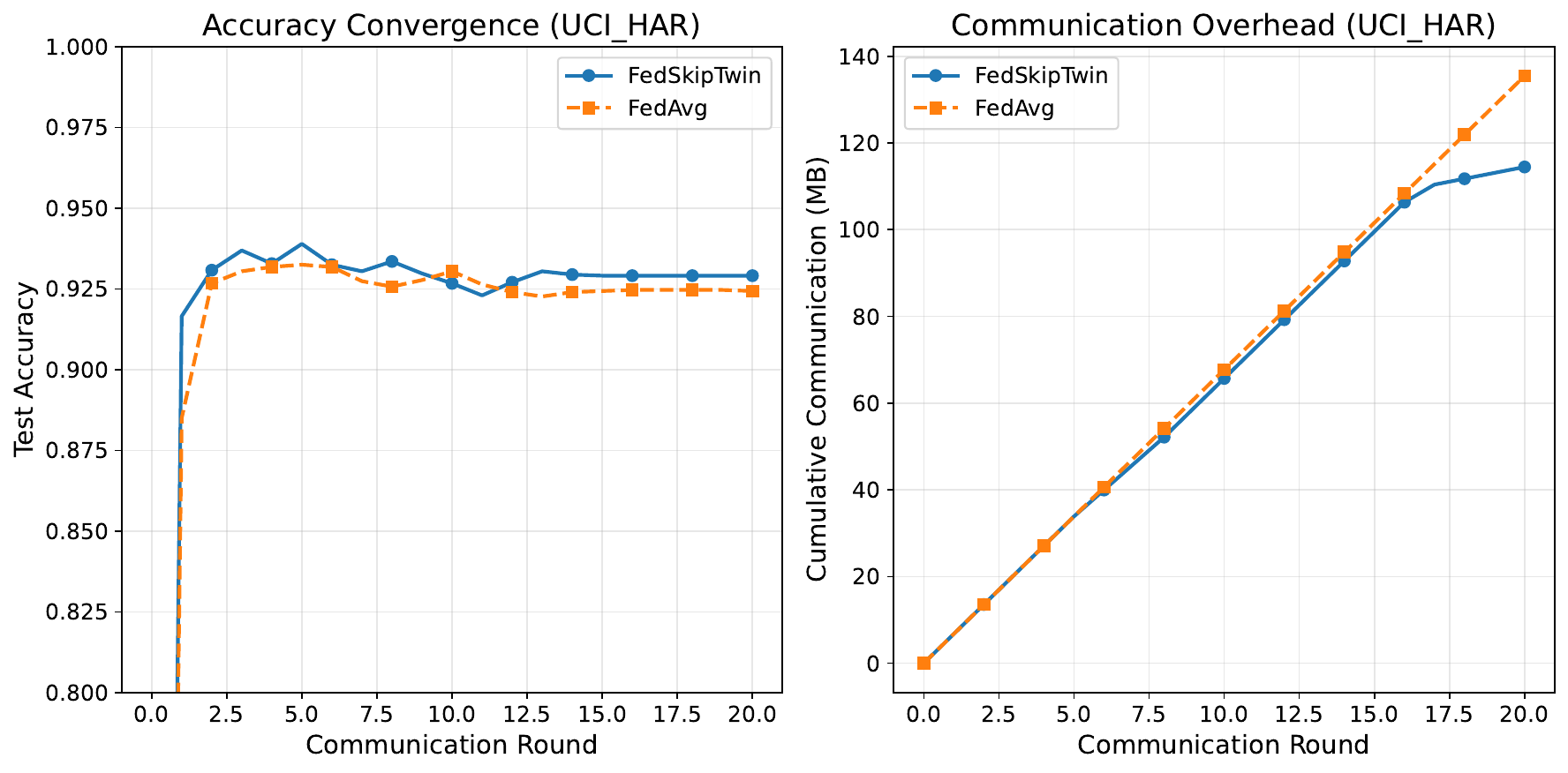}
\caption{Convergence comparison on the UCI-HAR dataset. \fedskiptwin\ achieves slightly higher accuracy with fewer communications.}
\label{fig:convergence_uci}
\end{figure}

\begin{figure}[!htbp]
\centering
\includegraphics[width=\linewidth]{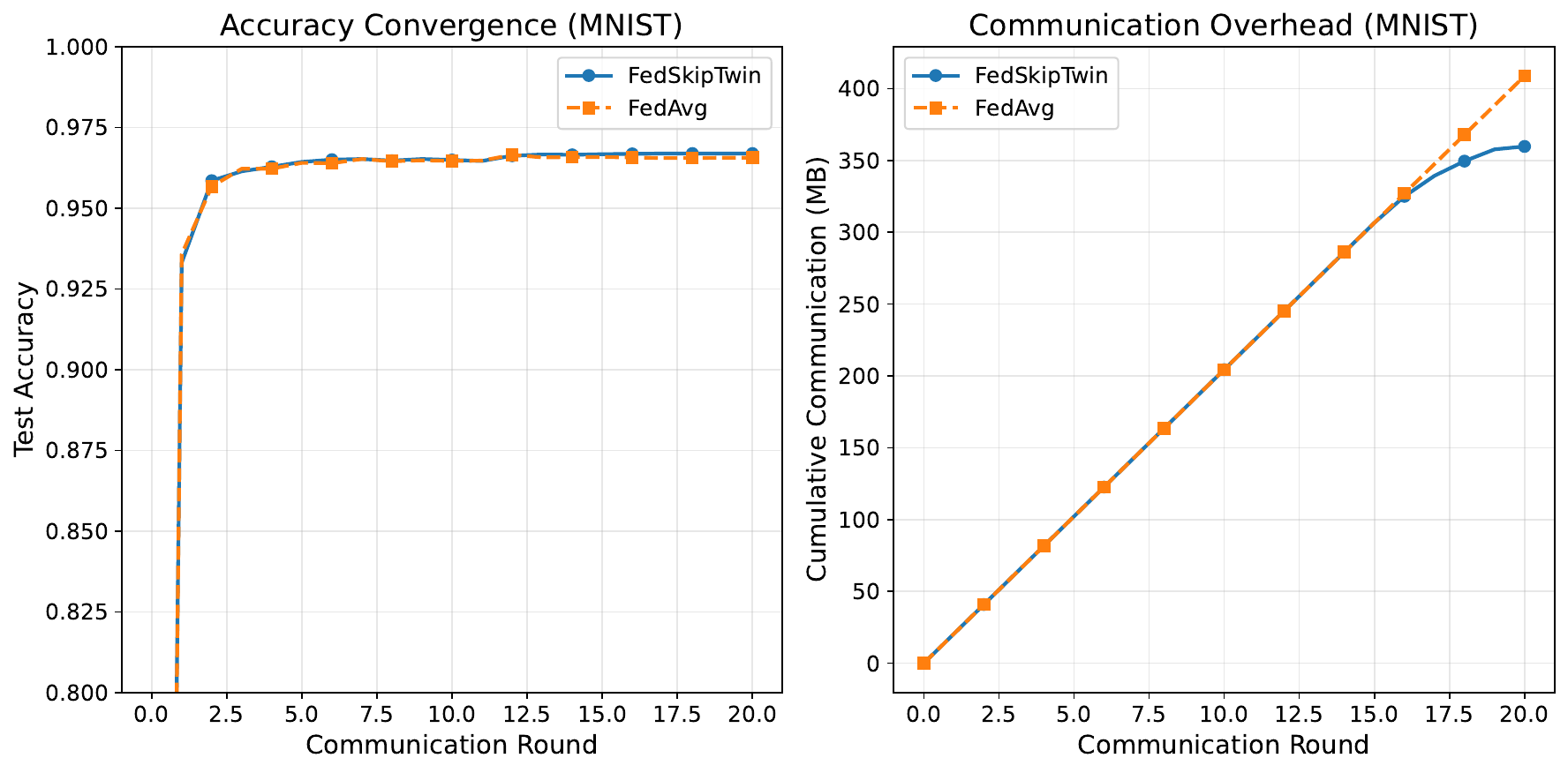}
\caption{Convergence comparison on the MNIST dataset. Both methods converge similarly, but \fedskiptwin\ uses less bandwidth.}
\label{fig:convergence_mnist}
\end{figure}

Figures \ref{fig:convergence_uci} and \ref{fig:convergence_mnist} plot the test accuracy over the 20 communication rounds. The convergence trajectory of \fedskiptwin\ closely tracks that of \fedavg\, demonstrating that skipping low-impact updates does not destabilize or slow down the learning process. The model reaches a high-quality solution at a rate comparable to the baseline, despite communicating less.

\begin{figure}[!htbp]
\centering
\includegraphics[width=\linewidth]{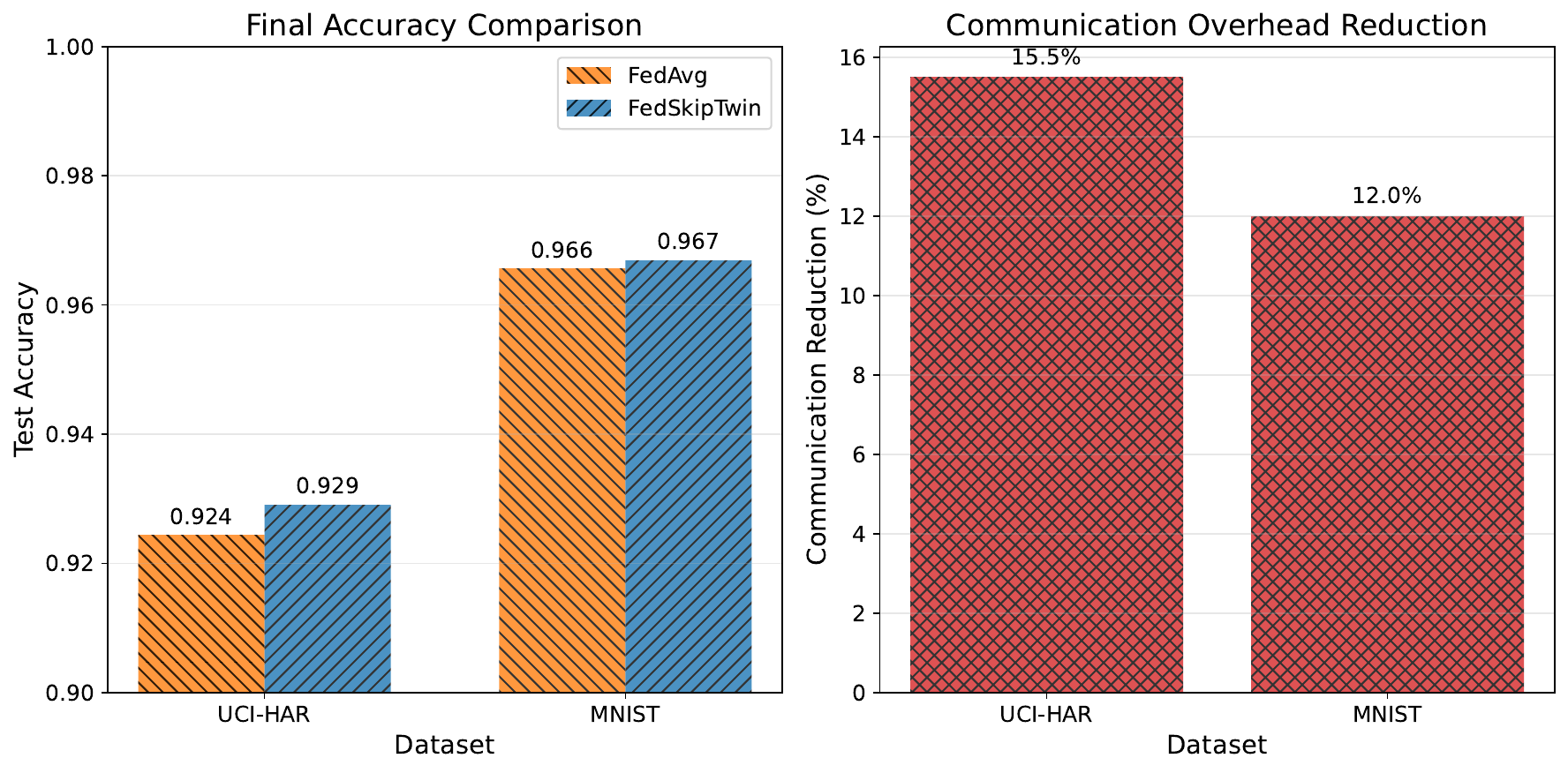}
\caption{Communication efficiency plot comparing final accuracy against total communication volume across datasets.}
\label{fig:efficiency}
\end{figure}

Figure~\ref{fig:efficiency} provides a consolidated view of the trade-off between communication cost and final accuracy. \fedskiptwin\ consistently occupies a more favorable position in the top-left quadrant, signifying higher accuracy for lower communication cost. This highlights its superior efficiency profile, making it a compelling choice for real-world FL on the edge.

\subsection{Skip Rate Dynamics}
\begin{figure}[h]
\centering
\includegraphics[width=\linewidth]{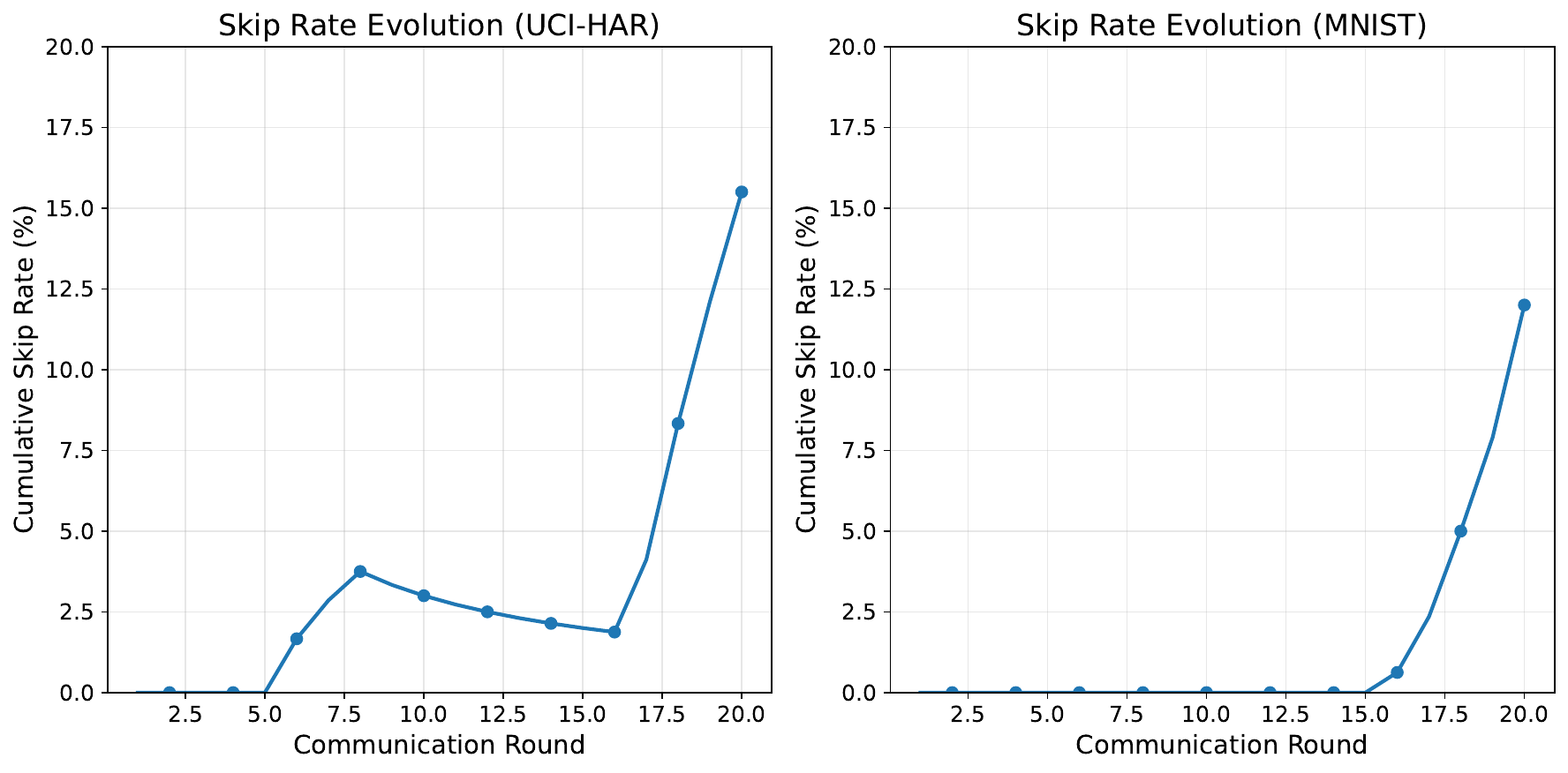}
\caption{Average client skip rate per round. The rate increases as twins become more accurate and the model converges.}
\label{fig:skiprate}
\end{figure}

Figure~\ref{fig:skiprate} illustrates the average client skip rate over the training duration. The rate is not static; it adapts as the digital twins learn and the global model evolves. Initially, the skip rate is low because the twins lack sufficient historical data to make confident predictions. As training progresses and the global model starts to converge, client updates naturally tend to become smaller in magnitude. The twins learn this pattern, leading to an increase in the skip rate in later rounds. This dynamic adjustment is key to balancing communication savings with model performance throughout the training lifecycle. The average skip rate across all rounds was \textbf{14.8\%} for UCI-HAR and \textbf{11.4\%} for MNIST.

\section{Discussion}
\label{sec:discussion}

\subsection{Key Insights}
The empirical success of \fedskiptwin\ is rooted in three key principles:
\begin{itemize}
    \item \textbf{Predictability of Update Significance}: The L2 norm of local gradients, which serves as a proxy for an update's importance, exhibits learnable temporal patterns. A simple LSTM is sufficient to capture these patterns effectively.
    \item \textbf{Server-Side Intelligence}: By hosting the digital twins on the server, our approach adds no computational or memory overhead to the resource-constrained client devices. The twin's overhead on the server is negligible.
    \item \textbf{Conservative, Uncertainty-Aware Skipping}: The dual-threshold mechanism, which considers both the predicted magnitude and the model's confidence, is crucial. It ensures that the system acts cautiously, preserving the integrity of the training process by avoiding risky skips.
\end{itemize}

\subsection{Limitations and Future Work}
While promising, this work has several limitations that open avenues for future research. First, the skip thresholds were determined by a simple grid search; an adaptive mechanism that dynamically adjusts these thresholds during training could yield better performance. Second, the "cold start" problem, where twins are inaccurate in early rounds due to sparse data, could be mitigated by using more sophisticated priors or transfer learning techniques for twin initialization. Finally, scaling this approach to thousands of clients will require investigating the performance of more advanced time-series models (e.g., Transformers) as twins and optimizing the server-side infrastructure.

\section{Conclusion}
\label{sec:conclusion}

This paper introduced \fedskiptwin, a novel federated learning algorithm designed to mitigate communication overhead by intelligently skipping client updates. By deploying lightweight, server-side digital twins that forecast the magnitude and uncertainty of future client gradients, our method preemptively avoids transmissions that are unlikely to contribute meaningfully to the global model. This server-centric intelligence imposes no extra burden on edge devices.

Our experiments on the UCI-HAR and MNIST datasets under a non-IID setting show that \fedskiptwin\ is highly effective. It reduced total communication volume by 12--15.5\% while achieving final model accuracy that was equal to or slightly better than the standard \fedavg\ algorithm. The results validate our central hypothesis: that the significance of client updates is predictable, and this predictability can be exploited to create more resource-aware federated learning systems. \fedskiptwin\ represents a practical step towards making FL more viable for deployment on large-scale networks of low-power edge devices.

\bibliographystyle{IEEEtran}
\bibliography{ref}

\end{document}